\documentclass[11pt]{article}
\usepackage{arxiv}
\usepackage{xurl}
\usepackage{tikz}
\usetikzlibrary{positioning,arrows.meta}
\usepackage{tabularx}
\hypersetup{
  pdftitle={Cascadia: Resident 975B MoE Inference on Eleven AI PCs},
  pdfauthor={Tate Berenbaum; Matias Parij; Muthaiah Venkatachalam}
}
\newcommand{\MTPFleet}{66.81}
\newcommand{\MTPDeployment}{64.36}

\newcommand{\SurveyDecode}{60.29}
\newcommand{\SurveyPhase}{46.87}
\newcommand{\SurveyTTFT}{6.05}
\newcommand{\SurveyPeakStreams}{88}

\newcommand{\ConcurrencyRows}{%
1 & 7.96 & 6.98 & 7.960 & 2.18 & 2.41 \\
2 & 4.53 & 4.37 & 2.267 & 2.06 & 3.03 \\
4 & 9.16 & 8.64 & 2.290 & 3.31 & 4.03 \\
6 & 13.89 & 12.80 & 2.315 & 4.20 & 5.49 \\
8 & 18.60 & 16.86 & 2.325 & 4.55 & 6.23 \\
11 & 23.28 & 20.86 & 2.116 & 5.47 & 7.64 \\
15 & 24.60 & 22.12 & 1.640 & 6.05 & 10.11 \\
22 & 33.87 & 29.26 & 1.539 & 9.71 & 14.51 \\
32 & 38.31 & 32.48 & 1.197 & 13.44 & 21.90 \\
48 & 45.33 & 37.68 & 0.944 & 18.85 & 31.55 \\
64 & 53.60 & 42.41 & 0.837 & 25.10 & 46.41 \\
88 & 60.29 & 46.87 & 0.685 & 34.61 & 64.75 \\
96 & 46.85 & 39.79 & 0.488 & 38.94 & 71.31 \\
128 & 50.10 & 41.61 & 0.391 & 55.11 & 108.73 \\
176 & 57.72 & 45.24 & 0.328 & 76.83 & 165.34 \\
}

\newcommand{\SpecRepeatRatio}{1.80}
\newcommand{\SpecRepeatPhaseRatio}{1.69}

\newcommand{\SpecGPUComparisonRows}{%
Explanation & 5.09 & 14.70 & 2.89$\times$ & +189.0\% \\
Code & 7.07 & 8.05 & 1.14$\times$ & +13.9\% \\
Arithmetic & 8.88 & 10.06 & 1.13$\times$ & +13.4\% \\
Story & 5.31 & 10.96 & 2.06$\times$ & +106.4\% \\
Tips & 3.55 & 11.66 & 3.28$\times$ & +227.9\% \\
Table & 3.76 & 8.66 & 2.30$\times$ & +130.2\% \\
Rewrite & 7.36 & 9.58 & 1.30$\times$ & +30.1\% \\
Facts & 6.12 & 10.12 & 1.66$\times$ & +65.5\% \\
Poem & 5.13 & 11.69 & 2.28$\times$ & +127.9\% \\
Instructions & 4.62 & 9.66 & 2.09$\times$ & +109.0\% \\
Translation & 8.00 & 9.23 & 1.15$\times$ & +15.4\% \\
True/false & 10.92 & 11.53 & 1.06$\times$ & +5.6\% \\
}
\newcommand{\SpecGPUMaxRatio}{3.28}
\newcommand{\SpecGPUExplainMedian}{11.27}
\newcommand{\SpecGPUExplainMax}{15.04}

\newcommand{\CtxRequests}{19}
\newcommand{\CtxNeedle}{19}

\newcommand{\CtxDecodeThirtyTwoK}{1.51}

\newcommand{\CtxDecodeSixtyFourK}{0.82}
\newcommand{\CtxPrefillRateSmall}{54}

\newcommand{\CtxLinearRate}{202}
\newcommand{\CtxQuadraticCoeff}{1.51}
\newcommand{\CtxQuadraticGflops}{11}
\newcommand{\CtxPredictHoursOneTwoEightK}{7.4}
\newcommand{\CtxPredictHoursTwoFiveSixK}{29}

\newcommand{\CtxCoresBusy}{1.00}
\newcommand{\CtxGpuBusyOneK}{56}
\newcommand{\CtxGpuBusySixtyFourK}{9}
\newcommand{\CtxFreeOneK}{6.98}
\newcommand{\CtxFreeSixtyFourK}{6.56}
\newcommand{\ProbeAttnSlope}{2.8}

\newcommand{\ProbeAttnSixtyFourK}{194}

\newcommand{\ProbeDecodeOneM}{2990}
\newcommand{\ProbeOneMNeedGB}{8.0}

\newcommand{\ProbeAttnGflopsSixtyFourK}{11}

\newcommand{\ContextRows}{1,041 & 3 & 22\,s & 46.4 & 4.73 (4.18--5.22) & 3/3 & 0.97 & 56\% & 6.98\\
2,011 & 3 & 37\,s & 53.7 & 5.82 (5.29--6.62) & 3/3 & 1.02 & 53\% & 6.93\\
3,991 & 3 & 75\,s & 52.9 & 3.91 (3.60--4.42) & 3/3 & 1.00 & 44\% & 6.99\\
7,889 & 3 & 2.9\,min & 45.7 & 3.32 (2.97--3.73) & 3/3 & 1.02 & 35\% & 6.95\\
15,897 & 3 & 7.8\,min & 33.8 & 2.71 (2.29--3.40) & 3/3 & 1.02 & 26\% & 6.90\\
31,614 & 2 & 27.2\,min & 19.3 & 1.51 (1.35--1.68) & 2/2 & 1.00 & 16\% & 6.76\\
64,292 & 2 & 109.7\,min & 9.8 & 0.82 (0.60--1.04) & 2/2 & 1.00 & 9\% & 6.56\\}

\title{Cascadia: Resident 975B MoE Inference\\on Eleven AI PCs}
\author{
  \begin{tabular}{@{}c@{\hspace{3em}}c@{}}
    Tate Berenbaum & Matias Parij \\
    Not Community Labs Inc. & Not Community Labs Inc. \\[0.75em]
    \multicolumn{2}{c}{Muthaiah Venkatachalam} \\
    \multicolumn{2}{c}{Intel Corporation}
  \end{tabular}
}
\date{October 2026}
\begin{document}
\maketitle

\begin{abstract}
Mixture-of-experts models make nearly trillion-parameter capacity accessible with sparse per-token computation, provided that the serving system can distribute the weights and coordinate their execution. We present Cascadia's resident execution of Inkling, a 975B-total/41B-active-parameter model, on eleven Intel Core Ultra X7 358H AI PCs, each with 64\,GB of memory, Arc B390 integrated graphics and gigabit Ethernet. We contribute a custom resident MoE engine that preserves Inkling's routing rules, constructs compressed graphs for OpenVINO's fused iGPU primitives, and coordinates FP16 expert computation with FP32 output restoration. The engine fits six consecutive decoder layers per machine and represents dense feed-forward blocks as all-active expert slices, reducing measured dense-layer call time from approximately 8.1 to 4.5\,ms. A streaming pipeline coordinates concurrent generation, while captured-state draft evaluation measures agreement with the deployed numerical path. Paired measurements at fifteen concurrency levels from 1 to 176 streams reach \SurveyDecode{} aggregate decode tokens/s at \SurveyPeakStreams{} streams, with \SurveyPhase{} tokens/s over the complete serving phases. At fifteen streams, median first-token latency is \SurveyTTFT{}\,s. Raising the context budget from the 1,024-position default, real prompts of 1k to 64k tokens recover the embedded code in all 19 measured answers, with first-token time growing as $aN+bN^2$ and decode latency growing approximately linearly, both bounded by a single-threaded CPU attention loop rather than by memory, which holds 512k positions per stream. Evaluation on captured fleet states separates the effects of vocabulary selection and weight quantization on draft agreement. Together, these contributions establish an execution and evaluation approach for large sparse models on distributed client systems with shared CPU--GPU memory.
\end{abstract}

\section{Introduction}
The sparse activation of mixture-of-experts (MoE) models changes the relationship between parameter capacity and computation. Inkling has 975B total parameters but activates 41B per token~\cite{inkling}. Serving such a model requires access to its expert weights even though each token selects only a subset. A collection of AI PCs offers a useful combination of distributed memory capacity, integrated matrix-compute accelerators and general-purpose CPU cores. Realizing that combination requires a serving architecture designed around shared CPU--GPU memory and sequential autoregressive dependencies.

Cascadia distributes Inkling's text decoder across eleven Intel Panther Lake machines. Each machine holds six consecutive decoder layers and executes its own portion of the model on the local CPU and integrated GPU. Compressed projections and fused expert operations execute on the GPU, while the CPU manages attention state, routing and stream coordination. Independent requests occupy different pipeline stages concurrently. A direct token-return connection closes the autoregressive loop, and bounded prefill windows allow prompt processing to progress through the same chain.

The research contribution is the design and evaluation of this integrated serving system at nearly trillion-parameter scale on 64\,GB client devices. The work develops three aspects that are closely coupled in this setting:

\begin{enumerate}
\item \textbf{A custom resident MoE engine for shared-memory accelerators.} We construct compressed graphs that preserve Inkling's routing rules while using OpenVINO's fused expert primitives, manage numerical range across the FP16/FP32 boundary, and coordinate layer residency with distributed stream state. Dense blocks use the same fused representation through eight all-active slices, taking approximately 4.5\,ms per measured call versus 8.1\,ms for three compressed matrix operations.
\item \textbf{A streaming pipeline for concurrent sparse-model service.} We integrate per-stream attention and convolution state, balanced admission, eight-row prefill windows and a direct sampled-token return path. A paired concurrency curve from 1 to 176 streams characterizes the resulting service. The highest measured paired mean occurs at 88 streams, eight per pipeline group: \SurveyDecode{} aggregate decode tokens/s and \SurveyPhase{} whole-phase tokens/s. Fifteen streams yield \SurveyTTFT{}\,s median first-token latency.
\item \textbf{Draft evaluation aligned with deployed inference.} We capture FP32 residuals together with the fleet's emitted token IDs and replay each draft module's temporal state. On 36 sequences, the resulting analysis separates vocabulary selection from additional weight quantization: restricting the head to 65,536 tokens changes first-draft agreement by 2.271 percentage points, while quantization at that vocabulary changes it by a further 0.175 percentage points.
\end{enumerate}

These contributions extend Cascadia's earlier pipeline-shard and architecture work~\cite{shards,architecture}. The new setting combines a much larger sparse model, compressed expert residency, dense/sparse operator unification and evaluation on distributed quantized states. We position the individual mechanisms relative to consumer-cluster inference, MoE execution and speculative decoding in Section~\ref{sec:related}.

\section{System architecture}
\label{sec:architecture}

\subsection{Panther Lake as a shared-memory inference platform}
Intel's Core Ultra X7 358H provides sixteen cores and threads: four performance cores, eight efficient cores and four low-power efficient cores. Its Arc B390 GPU has twelve Xe cores and uses the Xe3 graphics architecture. The processor supports LPDDR5X up to 9600\,MT/s and includes an NPU rated at 50 INT8 TOPS~\cite{intel358h,intelxe3}. Cascadia uses the CPU and integrated GPU for this deployment. CPU state processing and GPU matrix operations draw on the same local memory capacity and bandwidth.

\begin{table}[ht]
\centering\small
\begin{tabularx}{\linewidth}{lX}
\toprule
Component & Deployment configuration\\
\midrule
Fleet & Eleven Core Ultra X7 358H machines, each with nominal 64\,GB memory; 704\,GB nominal capacity across the fleet.\\
CPU and GPU & Sixteen logical CPUs and an Arc B390 integrated GPU per machine.\\
Usable memory & Operating-system reports of 62,757 or 62,615\,MiB per machine, approximately 61.15--61.29\,GiB.\\
Runtime & Linux, with recorded kernel \texttt{7.0.0-31-generic}; OpenVINO 2026.3.1 GPU execution and Rust serving/state management.\\
Network & Gigabit Ethernet; eight USB NCM interfaces and three PCIe-style interfaces.\\
Partition & Eleven pipeline roles, each owning six consecutive decoder layers.\\
Serving path & Text generation; deployment default of 1024 sequence positions, evaluated with short prompts and outputs.\\
\bottomrule
\end{tabularx}
\caption{The hardware and execution configuration used in the study. Vendor accelerator and memory-rate specifications describe hardware capability; serving performance is measured separately.}
\label{tab:hardware}
\end{table}

The fleet's aggregate capacity is distributed across independent address spaces. Model partitioning assigns weights and sequence state to each machine; network transfers carry the intermediate activations needed by the next stage. The system therefore exploits local shared memory within a machine and explicit message passing between machines.

\subsection{Model structure and partitioning}
Inkling's decoder has 66 layers. Its first two feed-forward blocks are dense, with intermediate width 24,576; the remaining 64 use routed and shared experts with intermediate width 3072. Each sparse layer selects six experts from a pool of 256 and also executes two shared experts. The hidden width is 6144. Attention combines sliding and global layers, learned relative-position bias and short convolutions~\cite{inklingcard}.

Pipeline role 0 owns layers 0--5, token embedding and request admission. Roles 1--9 each own six intermediate layers. Role 10 owns layers 60--65, final normalization, vocabulary projection and sampling. Forward messages contain FP32 residuals, stream-slot identifiers and positions. A width-6144 residual occupies 24,576 bytes per row before message metadata. The final role returns sampled token IDs directly to role 0, which schedules the next decoding step.

\begin{figure}[ht]
\centering
\resizebox{\linewidth}{!}{\begin{tikzpicture}[>=Latex,node distance=0.35cm,every node/.style={font=\small},box/.style={draw,rounded corners,align=center,minimum height=1.2cm,inner sep=6pt}]
\node[box,fill=blue!7] (first) {Role 0: layers 0--5\\admission, embedding\\per-stream scheduling};
\node[box,fill=blue!7,right=of first] (middle) {Roles 1--9\\six layers per machine\\independent stream state};
\node[box,fill=blue!7,right=of middle] (last) {Role 10: layers 60--65\\final norm, head, sampling};
\draw[->,thick] (first) -- (middle);
\draw[->,thick] (middle) -- (last);
\draw[->,thick] (last.south) -- ++(0,-0.45) -| node[pos=.25,below]{direct token-ID return closes the autoregressive loop} (first.south);
\node[box,fill=orange!8,below=1.15cm of middle,minimum width=12cm] {Within each machine: iGPU compressed projections and fused experts\\CPU attention, routing, KV/convolution state and coordination; shared local memory};
\end{tikzpicture}}
\caption{Cascadia's resident streaming pipeline. Different requests can occupy different stages concurrently; consecutive ordinary tokens of one request depend on the sampled-token return.}
\label{fig:architecture}
\end{figure}
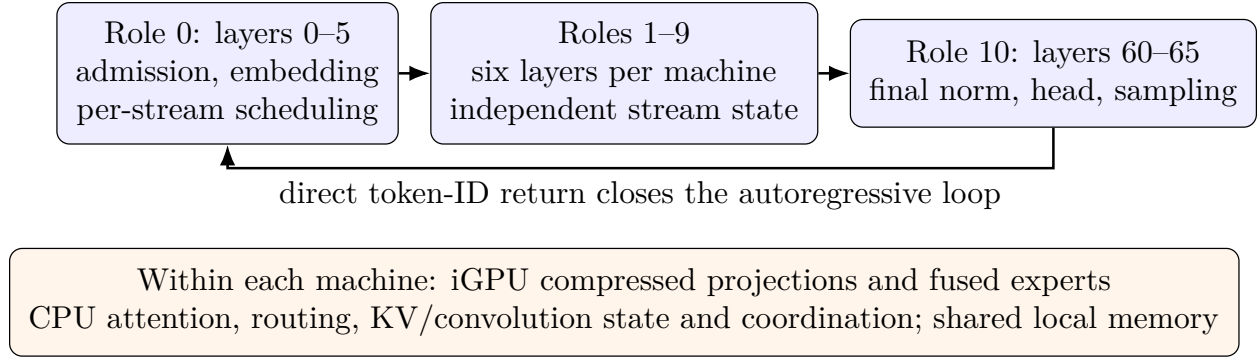

\section{A custom resident MoE engine for integrated GPUs}
\label{sec:resident}

\subsection{Model-specific graph construction and routing}
Cascadia's engine connects Inkling's model semantics to OpenVINO's compressed fused-MoE primitive and GPU kernel implementations~\cite{ovmoe}. Cascadia supplies the exporter, routing interface, numerical range management and resident layer runtime. OpenVINO supplies graph recognition, lowering and device execution. This division lets the engine use established accelerated operations while retaining the model's routing and state behavior.

The exporter builds one graph per MoE layer from the existing group-32 INT4 expert bins. It arranges weights in expert-major order, preserves their packed nibbles, and represents group scales in FP16. A tiled three-matrix-operation graph expresses gate/up projections, SwiGLU, down projection and weighted reduction in the pattern recognized by OpenVINO's MoE lowering pass~\cite{ovmoelowering}. This tiled algebra is a compiler representation; the fused path executes the selected expert work.

Routing remains in Rust. The gate computes Inkling's sigmoid scores, bias-adjusted top-six selection, selected/shared normalization and model-specific scale factors. Its results enter the graph as expert IDs and routing weights. The two shared experts occupy IDs 256 and 257 and are always included, so each row supplies eight selections through the same interface. This preserves Inkling's routing rules while allowing the GPU backend to fuse the expert computation.

\begin{figure}[ht]
\centering
\resizebox{\linewidth}{!}{\begin{tikzpicture}[>=Latex,every node/.style={font=\footnotesize},box/.style={draw,rounded corners,align=center,text width=4.3cm,minimum height=1.55cm,inner sep=5pt}]
\node[box,fill=blue!7] (export) at (0,0) {Cascadia exporter\\expert-major INT4 graphs\\and layer scale metadata};
\node[box,fill=orange!10] (compile) at (5.2,0) {OpenVINO compiler\\MoE pattern recognition\\and compressed fusion};
\node[box,fill=blue!7] (resident) at (10.4,0) {Cascadia layer runtime\\resident compiled models\\and reusable requests};
\draw[->,thick] (export) -- (compile);
\draw[->,thick] (compile) -- (resident);
\node[box,fill=blue!7] (route) at (0,-2.5) {Cascadia host path\\routing IDs, scaled weights\\and frame rows};
\node[box,fill=orange!10] (gpu) at (5.2,-2.5) {OpenVINO iGPU path\\compressed expert execution\\with FP16 arithmetic};
\node[box,fill=blue!7] (restore) at (10.4,-2.5) {Cascadia host path\\FP32 output restoration\\convolution and residual};
\draw[->,thick] (route) -- (gpu);
\draw[->,thick] (gpu) -- (restore);
\draw[->,dashed] (resident.south) -- ++(0,-.45) -| (gpu.north);
\end{tikzpicture}}
\caption{The custom engine and its OpenVINO execution boundary. The upper path prepares resident layer models; the lower path executes each frame. Cascadia supplies the model-specific representation and stateful runtime around OpenVINO's compiler and GPU primitives.}
\label{fig:engine}
\end{figure}
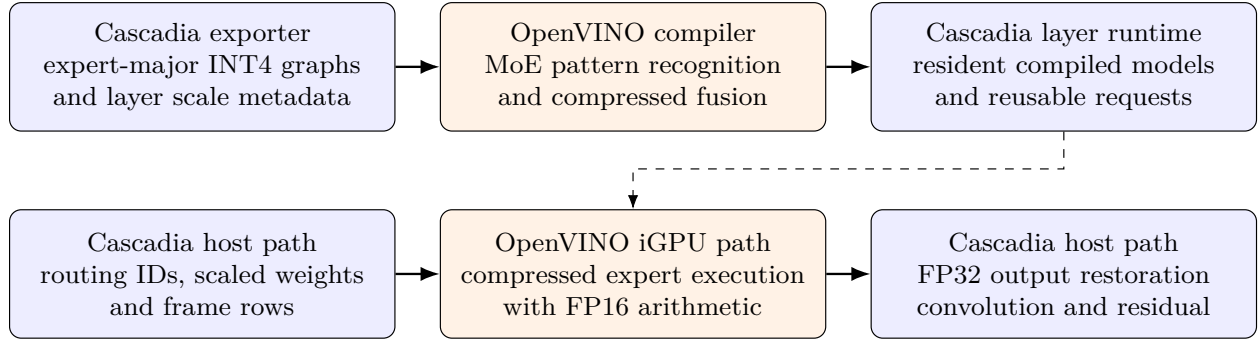

\subsection{Residency and the layer execution contract}
Expert weights use four-bit compression; attention projections and the output head use eight-bit weights. CPU routing, attention and convolution history share local memory with GPU weights and working storage. The configuration budgets approximately 52\,GiB for GPU pages per machine, with a larger allocation on the final role, and avoids a second large resident CPU expert cache in normal operation.

Cascadia's C++ bridge materializes graph constants before compilation for the resident path. The Rust runtime retains one compiled model and reusable inference request per layer. A call supplies a frame's rows, expert IDs and weights; the runtime selects a supported row shape, pads extra rows with zero routing weights, invokes the compiled model and restores the real row count on return. Host computation then advances the stream's convolution and residual state. Device-call counters and profiles make the execution placement observable. The allocation and scheduling unit is the complete layer shard, including weights, sequence state and host working memory.

\subsection{Numerical range across the FP16/FP32 boundary}
The engine manages two sources of magnitude separately: the expert output and its routing coefficient. For layer $\ell$, the exporter can attenuate up-projection scales by $A_\ell=2^{n_\ell}$ and record the exponent with the graph. Since the up branch enters SwiGLU multiplicatively, this scales the expert output $E_{\ell e}(x)$ by $1/A_\ell$ in real arithmetic. The recorded layer-8 configuration uses $A_8=16$.

For row $i$, with $e$ ranging over its eight selected experts, the runtime chooses
\begin{equation}
 F_i=2^{\lceil\log_2\max(1,\sum_e |w_{ie}|)\rceil},\qquad
 \widetilde y_i=\sum_e\frac{w_{ie}}{F_i}\frac{E_{\ell e}(x_i)}{A_\ell},\qquad
 y_i=A_\ell F_i\widetilde y_i.
 \label{eq:range}
\end{equation}
The GPU receives the normalized routing weights, whose absolute values sum to at most one. In real arithmetic, their weighted combination is bounded componentwise by the largest absolute attenuated expert value. The host restores both factors in FP32 before state updates. Layer metadata binds the export-time attenuation to the runtime compensation.

Equation~\ref{eq:range} preserves the real-arithmetic function. Powers of two change exponents exactly when representable; FP16 conversion, subnormal scales and reduction rounding still determine finite-precision agreement. Equivalent rescaling is established numerical practice: SmoothQuant, for example, transfers activation magnitude into weights for W8A8 quantization~\cite{smoothquant}. Our implementation applies separate layer and row factors to preserve the fused FP16 expert path and the FP32 residual interface. This is an explicit engine design and implementation contribution, evaluated together with resident serving.

\subsection{A common fused representation for dense and sparse blocks}
The two dense feed-forward blocks can use the same fused-expert operator family as the sparse layers. Partitioning the intermediate dimension into eight slices of width 3072 gives
\begin{equation}
 W_d[\operatorname{SiLU}(W_gx)\odot W_ux]
 =\sum_{j=1}^{8}W_{d,j}[\operatorname{SiLU}(W_{g,j}x)\odot W_{u,j}x].
 \label{eq:dense}
\end{equation}
Every slice is active with unit weight. Gate/up rows and down-projection columns are sliced at existing quantization-group boundaries, preserving packed weights without requantization. The dense block thus matches the backend's expert width without introducing a learned router or dropping intermediate activations. The algebra follows established dense-to-expert decompositions~\cite{moefication,mlpmoe}; our implementation maps that representation onto the compressed fused operator used by the resident Inkling shard.

\begin{table}[ht]
\centering\small
\begin{tabular}{lrrrr}
\toprule
Dense layer & Three matrix calls & Fused slices & Time reduction & Relative difference\\
\midrule
Layer 0 & 8.15\,ms & 4.51\,ms & 44.7\% & $5.7\times10^{-4}$\\
Layer 1 & 8.11\,ms & 4.45\,ms & 45.1\% & $5.9\times10^{-4}$\\
\bottomrule
\end{tabular}
\caption{One-row load checks for the dense blocks. Relative difference compares the two FP16 execution paths. Equation~\ref{eq:dense} preserves the real-arithmetic function; the measured differences reflect finite-precision execution.}
\label{tab:dense}
\end{table}

\begin{figure}[ht]
\centering\includegraphics[width=.88\linewidth]{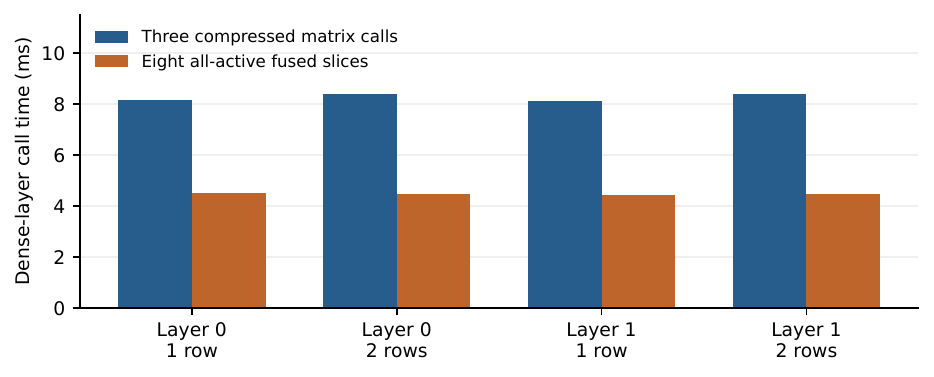}
\caption{Dense-block execution using three compressed matrix operations or eight all-active fused slices. The operator improvement holds for the measured one- and two-row calls.}
\label{fig:dense}
\end{figure}

The two-row measurements are 8.41\,ms per layer with the matrix-call path and 4.47/4.48\,ms with fused slices. At the stage level, role 0's observed computation changes from 51.9 to 43.7\,ms per frame. These are operator and stage measurements; the complete serving rates reported in Section~\ref{sec:performance} characterize the distributed system separately.

\subsection{Weight traffic and the role of concurrent requests}
Using the exported compressed sizes, an unshared full-model token accounts for approximately 16.76\,GB of active expert weights, 8.59\,GB of attention projections, 0.52\,GB of dense weights and 1.24\,GB of output-head weights: about 27.1\,GB in total. These are logical weight bytes, including expert scale/zero-point overhead, rather than a count of physical DRAM transactions. Caching and reuse across rows determine how those bytes translate into memory traffic~\cite{roofline}.

The layout makes two forms of parallelism explicit. Within a stage, a multi-row frame shares projection work and allows CPU attention to run across independent streams. Across stages, different frames use different memory buses concurrently. Consequently, aggregate service is a property of the distribution of requests over the pipeline as well as of the local kernels. This motivates the stream and admission design below.

\section{Streaming service across the fleet}
\label{sec:streaming}

\subsection{Stateful streams and direct token return}
Each request occupies a stream slot with its own position, KV cache and convolution state. Frames group rows from independent slots, and each stage applies its local layers before forwarding the residuals. Role 0 admits work into the least populated group and advances the engine in short steps, allowing request submission to proceed while other frames are circulating.

The direct final-to-first connection carries sampled token IDs without requiring a reverse traversal through the intermediate workers. Receipt of an ID makes the next token of that stream schedulable. This explicit return path lets ingress interleave admission, forward progress and completion handling within the same serving loop. A full-chain readiness probe establishes downstream readiness before admission begins.

\subsection{Prefill windows and parallel attention}
Prompts enter as windows of at most eight rows. Consecutive windows can occupy different pipeline stages, and admission can proceed while ingress waits for token replies. The window size bounds the prefill work presented to a stage at one time. During decoding, independent rows provide parallel CPU work for attention even while the integrated GPU handles compressed projections and experts.

Chunked prefill and iteration-level scheduling have established precedents~\cite{orca,sarathi,sarathiserve}. The contribution here is their realization within a resident sparse-model pipeline whose stages maintain both KV and convolution state, share memory between CPU and GPU, and communicate over gigabit Ethernet. Prefill, decode, row-parallel attention and direct token return are coordinated in one runtime.

\subsection{Speculative decoding scenarios and measured gains}
\label{sec:speculation}
A lone active request ordinarily waits for a token to traverse all eleven stages before its next token can enter the pipeline. Cascadia overlaps this dependency by injecting proposed future input tokens behind the confirmed input, as separate frames in the same pipeline. Each returning target-model reply verifies the input of the following frame. A matching proposal retains computation already in flight; a mismatch discards that frame and its speculative successors, drains their replies and rewinds every stage's KV and short-convolution state to the confirmed prefix. Generation resumes with the target's token. Emitted tokens therefore come from target replies, with agreement assessed under the greedy serving configuration. This asynchronous implementation extends established speculative decoding and Cascadia's earlier pipelined speculation~\cite{specdecode,shards} to Inkling's stateful layers.

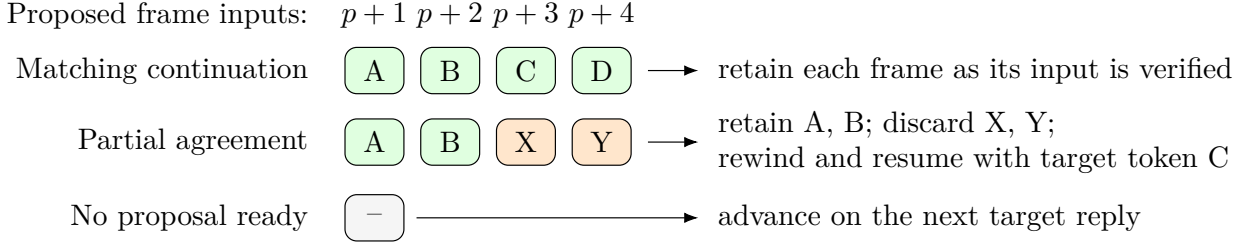
\begin{figure}[ht]
\centering
\resizebox{\linewidth}{!}{\begin{tikzpicture}[>=Latex,every node/.style={font=\small},token/.style={draw,rounded corners,minimum width=.7cm,minimum height=.55cm}]
\node[anchor=east] at (0,.65) {Proposed frame inputs:};
\foreach \x/\p in {0.65/1,1.55/2,2.45/3,3.35/4}
  \node at (\x,.65) {$p+\p$};
\node[anchor=east] at (0,0) {Matching continuation};
\foreach \x/\t in {0.65/A,1.55/B,2.45/C,3.35/D}
  \node[token,fill=green!12] at (\x,0) {\t};
\draw[->] (3.9,0) -- (4.5,0);
\node[anchor=west] at (4.6,0) {retain each frame as its input is verified};
\node[anchor=east] at (0,-.85) {Partial agreement};
\foreach \x/\t in {0.65/A,1.55/B}
  \node[token,fill=green!12] at (\x,-.85) {\t};
\foreach \x/\t in {2.45/X,3.35/Y}
  \node[token,fill=orange!20] at (\x,-.85) {\t};
\draw[->] (3.9,-.85) -- (4.5,-.85);
\node[anchor=west,align=left] at (4.6,-.85) {retain A, B; discard X, Y;\\rewind and resume with target token C};
\node[anchor=east] at (0,-1.75) {No proposal ready};
\node[token,fill=gray!8] at (.65,-1.75) {--};
\draw[->] (1.15,-1.75) -- (4.5,-1.75);
\node[anchor=west] at (4.6,-1.75) {advance on the next target reply};
\end{tikzpicture}}
\caption{Speculative pipeline scenarios, assuming the target continuation is A, B, C, D. Letters denote symbolic token IDs. Frames enter successively and their inputs are verified by preceding target replies; proposed tokens are never emitted without verification.}
\label{fig:speculation}
\end{figure}

\paragraph{Proposal scenarios.}
Copying and recurring phrases can obtain proposals directly from request-local history or a persisted phrase table. The hybrid proposer favors a request-local four-token suffix match or a shared three-token context seen at least twice with at least 90\% next-token agreement. For other contexts, a rank-0 CPU draft model, Qwen3-0.6B in Q4\_K\_M format, can supply continuations. Its text is encoded with Inkling's tokenizer before injection. Figure~\ref{fig:speculation} illustrates complete agreement, partial agreement and ordinary progress when no proposal is ready. A mismatch preserves the verified prefix while restoring both forms of temporal state. When another request awaits admission, the runtime returns to its ordinary concurrent-stream path.

\paragraph{Why correct proposals improve latency.}
Let $L$ denote full-pipeline latency, $T$ the inter-frame interval and $a$ the fraction of output positions with correct inputs already in flight. A first-order model gives throughput $Q_{\mathrm{spec}}$ and speedup $S$ over ordinary decoding:
\begin{equation}
Q_{\mathrm{spec}}\approx\frac{1}{aT+(1-a)L},\qquad
S\approx\frac{L}{aT+(1-a)L}.
\label{eq:speculation}
\end{equation}
Correct proposals can replace full round trips with shorter inter-frame intervals. Fused iGPU execution reduces the stage-computation cost, while effective proposals overlap successive positions across the eleven devices. The illustration omits additional drafting, queueing and rewind costs; $a$ measures useful in-flight work, rather than conditional draft acceptance or offline MTP agreement.

\paragraph{Measured gains with the fused iGPU engine.}
Table~\ref{tab:speculation} reports all twelve paired single-stream prompts from the finalized GPU deployment. Target experts, dense layers, attention projections and the output head use the iGPU engine; the small external drafter runs on the CPU. Each prompt produces the same complete 128-token output in both passes, with the serving binary and model fixed. The largest observed per-request rate ratio is \SpecGPUMaxRatio{}$\times$ for tips (3.55 to 11.66 tokens/s); explanation reaches 14.70 tokens/s from 5.09 (2.89$\times$), and table generation rises from 3.76 to 8.66 (2.30$\times$). Showing every pair makes the workload dependence explicit.

\begin{table}[ht]
\centering\small
\begin{tabular}{lrrrr}
\toprule
Prompt family & First pass & Repeated pass & Rate ratio & Gain\\
\midrule
\SpecGPUComparisonRows
\bottomrule
\end{tabular}
\caption{Fused-iGPU target inference: first/repeated decode tokens/s for one prompt from each family, 128 output tokens per request. All twelve pairs have identical prompts and complete output text. Speculation is enabled in both passes; these are observed repeated-prompt gains with accumulated history and differing capture state.}
\label{tab:speculation}
\end{table}

Across the twelve serial requests, decode throughput rises from 5.68 to 10.24 tokens/s (\SpecRepeatRatio{}$\times$); whole-phase throughput rises from 5.19 to 8.77 (\SpecRepeatPhaseRatio{}$\times$). These phase rates account for the requests' actual durations rather than averaging their individual speedups. Both passes enable speculation and phrase learning. The first study pass already has prior history; further history accumulates before the repeat, and residual-capture writes are disabled between passes. The comparison characterizes the GPU system's observed repeated-prompt service, with one observation per prompt per pass, rather than isolating a speculation on/off effect.

\paragraph{Reusable history and structured GPU workloads.}
An additional fused-iGPU comparison illustrates reuse of learned phrases after transferring history to the current entry role. Merging the phrase tables increases retained contexts from 24,453 to 194,299. For six previously seen 128-token prompts measured before and after the merge, code rises from 5.23 to 10.60 tokens/s (2.03$\times$) and true/false from 4.91 to 10.77 (2.19$\times$). The binary remains unchanged; repeated requests also train the table, and capture is disabled with the transfer. These observations establish reusable history within GPU serving, with those timing influences included.

The hybrid proposer also serves structured requests with related-family history: retained GPU examples reach 6.60 tokens/s for translation, 8.19 for arithmetic and 10.95 for true/false completion. A separate finalized explanation-family phase contains three 128-token single-stream requests with median \SpecGPUExplainMedian{} tokens/s and fastest rate \SpecGPUExplainMax{} tokens/s. These are absolute request rates on the stated prompts. The GPU comparisons concern lone-stream speculation; the peak concurrent throughput in Section~\ref{sec:performance} measures a separate serving regime. The artifact retains the earlier CPU-expert on/off checks as supporting protocol evidence.

\section{Evaluation of concurrent serving}
\label{sec:performance}

\subsection{Measurement protocol}
The finalized survey holds the model and serving binary fixed: group-32 INT4 experts, INT8 attention/head weights and the streaming configuration described above. Thirty mixed-workload phases cover fifteen concurrency levels from 1 to 176, with two phases per level and twelve deterministic prompt families. Each request uses greedy generation and emits 128 tokens. Requests within a cohort arrive approximately 15\,ms apart; at low concurrency, successive complete cohorts supply at least twelve requests per phase. The complete survey also retains three explanation-family phases, for 33 measured phases, 1,592 requests and 203,776 generated tokens. Table~\ref{tab:service} uses the thirty mixed-workload phases.

The paired passes reuse the same prompt templates. The main sweep proceeds upward and then downward in concurrency, with the 88-stream refinement measured as an additional pair. Phrase learning remains enabled throughout. Residual-capture writes are enabled at the ascending 1--15-stream points and disabled thereafter, including the repeated pass. The pairs therefore describe the serving system with its accumulated history and recorded instrumentation state. Within high-concurrency cohorts, the finite prompt generator also produces some repeated inputs: the 88-stream phase contains 79 distinct prompts. The runtime release and binary hash, phase records and token traces accompany the artifact.

For cohort $j$, let $f_i$ and $l_i$ be the first and last token-event times of each request. The common decode interval is $(a_j,b_j]$, where $a_j=\max_i f_i$ and $b_j=\min_i l_i$. Let $u_j$ count tokens emitted by all requests during that interval. We report
\begin{equation}
 Q_{\rm decode}=\frac{\sum_j u_j}{\sum_j(b_j-a_j)},\qquad
 Q_{\rm phase}=\frac{\sum_i n_i}{t_{\rm end}-t_{\rm start}}.
 \label{eq:metrics}
\end{equation}
$Q_{\rm decode}$ measures aggregate throughput while every request in a cohort is decoding. $Q_{\rm phase}$ includes admission, prefill, queueing and drain. Dividing $Q_{\rm decode}$ by concurrency gives the mean contribution per stream over these common intervals. Table and figure rates are arithmetic means of the two phase rates; TTFT quantiles pool the requests from both phases. The observed range of two phases is descriptive, rather than a confidence interval.

The client counts token-event multiplicities and checks them against each request's final API token usage. The retained traces independently reconstruct all common intervals and whole-phase rates. TTFT refers to the first emitted model-token event, including reasoning and structural tokens. The collector checks server counters around each phase to isolate measured traffic. Software profiles supply frame-weighted stage timings, with physical device identity kept separate from pipeline role. Short serial/concurrent prefix checks and twelve known-answer questions provide functional validation; they are not a general model-quality benchmark.

\subsection{Concurrency, throughput and first-token latency}
\begin{table}[ht]
\centering\small
\begin{tabular}{rrrrrr}
\toprule
Streams & $Q_{\rm decode}$ & $Q_{\rm phase}$ & $Q_{\rm decode}/C$ & TTFT median & TTFT p95\\
\midrule
\ConcurrencyRows
\bottomrule
\end{tabular}
\caption{Finalized mixed-workload concurrency curve. Every request produces 128 tokens. Rates are tokens/s; TTFT is in seconds. $C$ is concurrent streams, rather than total requests at the low-concurrency points. Rates average two phases; TTFT quantiles pool their requests.}
\label{tab:service}
\end{table}

\begin{figure}[ht]
\centering\includegraphics[width=\linewidth]{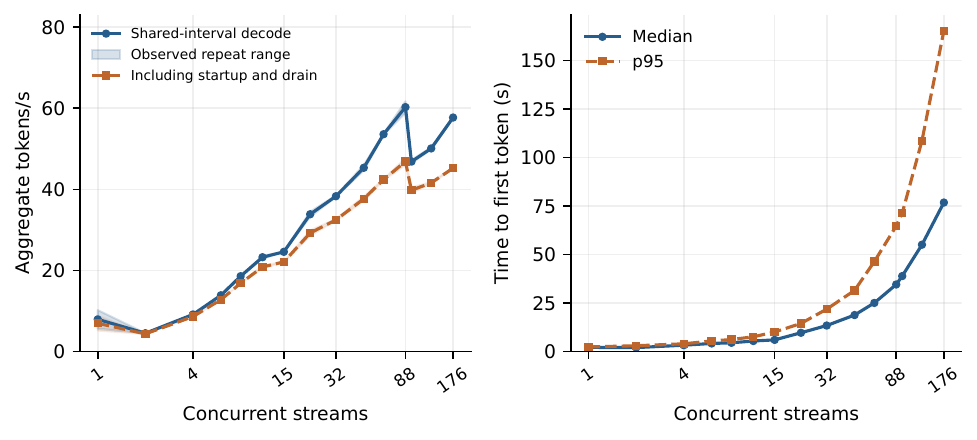}
\caption{Paired measurements across fifteen concurrency levels. Left: common-interval decode throughput and whole-phase throughput, with shading showing the two observed phase rates. Right: pooled request TTFT median and 95th percentile. The 88-stream point is the highest paired mean on the measured grid.}
\label{fig:concurrency}
\end{figure}

At 88 streams, the two phases each generate 11,264 tokens. Common-interval decode rates are 58.86 and 61.71 tokens/s; whole-phase rates are 46.03 and 47.72 tokens/s. Their paired means, \SurveyDecode{} and \SurveyPhase{} tokens/s, are the highest on the evaluated concurrency grid for both measures. This operating point assigns eight streams to each of eleven pipeline groups. At 176 streams the fleet sustains 57.72 aggregate decode tokens/s and 45.24 whole-phase tokens/s, while fifteen streams provide 24.60 and 22.12 tokens/s with \SurveyTTFT{}\,s median TTFT.

The curve exposes a deployment-specific relationship between concurrency and the distributed execution schedule. The 88-stream alignment coincides with the highest observed throughput, and further concurrency redistributes work into larger per-group row counts. Throughput is nonmonotonic across that transition, while first-token latency grows as more requests enter the chain. These observations identify a useful measured operating point for this resident pipeline; they do not by themselves isolate the effects of row padding, expert routing and scheduling. The contribution is an empirical service envelope that connects the implemented architecture to explicit throughput and latency choices.

\subsection{Effect of windowed admission}
The reference fifteen-request burst configuration records median TTFTs of 31.52 and 31.23\,s and whole-phase rates of 15.678 and 17.005 tokens/s. The windowed configuration records 6.91\,s median TTFT and 22.521 whole-phase tokens/s. The median latency reduction is approximately 4.5-fold across these observed configurations.

\begin{figure}[ht]
\centering\includegraphics[width=\linewidth]{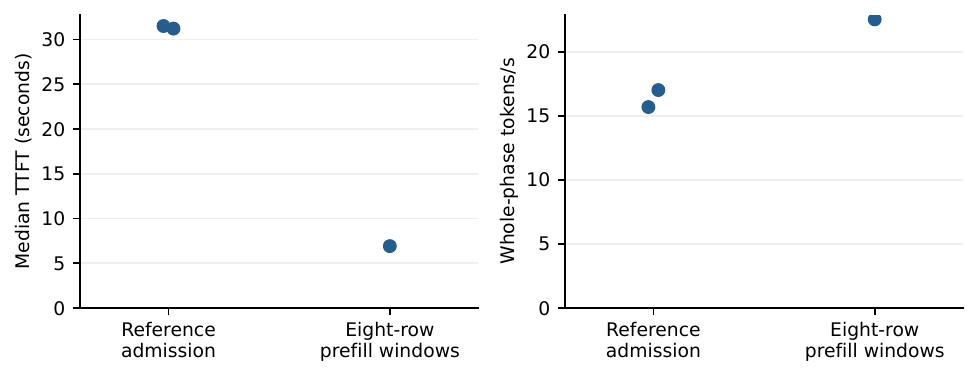}
\caption{Fifteen-request bursts with an output cap of 128: reference admission and eight-row prefill windows. Each point is a retained phase. The comparison reports observed configurations with different prompt tags.}
\label{fig:prefill}
\end{figure}

This supporting configuration comparison uses distinct prompt tags and precedes the finalized fixed-binary survey. It characterizes the admission design; Table~\ref{tab:service} characterizes the resulting deployment across concurrency levels. With one-second arrival spacing, the windowed configuration also records 2.48\,s median TTFT for fifteen requests capped at 96 output tokens.

\subsection{Context length}
\label{sec:context}
The serving configuration above holds 1,024 sequence positions per stream, the fleet default (Section~\ref{sec:architecture}). Inkling supports up to a million positions, so we measured what the deployment does when the budget is raised. Two things were measured. A \emph{probe} on every machine, run at worker start with the position budget set to $2^{20}$, filled one stream slot's caches to $N$ positions with synthetic rows (every byte a real sequence writes), decoded one row at that position several times, reported memory and time, and freed the pages; sizes that would not fit in the machine's free memory with a 3\,GiB margin were reported and not attempted. Then \emph{real prompts}, natural text of about $N$ tokens with a sentence naming a code near the start and a question asking for it at the end, were served one stream at a time through the API with 96 output tokens, with three repeats up to 16k tokens, two up to 64k and one above, while every machine's load was sampled every five seconds. The test used the deployed engine with two additions it required: a prompt travels down the pipeline one window per engine step, and a step that only sends a window used to return nothing, so the runner's watchdog closed any prompt longer than about twenty-two windows; the feed now reports progress per window. The prompt window's device shapes are compiled at load rather than on the first long prompt. Prompts traveled as 64-row windows; the production 8-row window prefills at the same rate.

\begin{table}[ht]
\centering\small
\begin{tabular}{rrrrrrrrr}
\toprule
Context & $n$ & First token & Prefill tok/s & Decode tok/s (range) & Code & Cores & iGPU & Free\\
\midrule
\ContextRows
\bottomrule
\end{tabular}
\caption{One stream at increasing context. Context is the API's prompt token count; $n$ repeats; first token is the mean (repeats agree to the second); decode is the mean and range over repeats; Code counts answers that contain the code stated near the start of the prompt; Cores is the mean number of busy CPU cores per machine during the request; iGPU its mean busy share; Free the least free memory of any machine in GiB. The 128k prompt did not produce its first token within the six-hour request cap.}
\label{tab:context}
\end{table}

\begin{figure}[ht]
\centering\includegraphics[width=\linewidth]{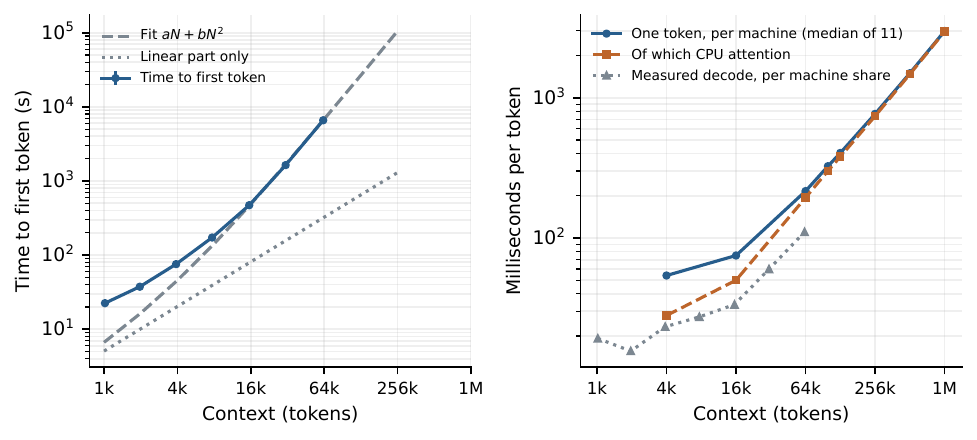}
\caption{Left: time to first token against context for one stream, with the fitted $aN+bN^2$ and its linear part. Right: the per-machine cost of one decode token from the probe (median of the eleven machines; 1M on the one machine that fits it) and the CPU attention within it, next to the measured single-stream decode divided by eleven; the measured points sit below the probe at small contexts because a lone stream's accepted draft guesses add tokens without a full pass.}
\label{fig:context}
\end{figure}

Table~\ref{tab:context} and Figure~\ref{fig:context} report the results. The code was found in \CtxNeedle{} of \CtxRequests{} answers: the global layers' attention over the whole context works at 64k, well beyond the 1,024 positions of the relative-position bias. Memory was not the limit at these sizes: the 64k context cost 0.5\,GiB per machine (8\,KiB per position on the one global layer per shard, FP32), from \CtxFreeOneK{} to \CtxFreeSixtyFourK{}\,GiB free. The probe admits 512k positions on every machine while retaining its 3\,GiB free-memory margin. At 1M positions, the cache requires \ProbeOneMNeedGB{}\,GiB per machine; only the machine serving rank 0 satisfies the margin, and it runs the probe at \ProbeDecodeOneM{}\,ms per token. These are memory-capacity and synthetic decode observations, distinct from end-to-end prompt serving.

Time is the limit. The first-token time fits
\begin{equation}
 T_{\rm prefill}(N)=\frac{N}{\CtxLinearRate{}\ \text{tokens/s}}+\CtxQuadraticCoeff{}\times10^{-6}\,\text{s}\times N^{2},
 \label{eq:prefill}
\end{equation}
which predicts \CtxPredictHoursOneTwoEightK{}\,h at 128k, consistent with the request that did not finish, and \CtxPredictHoursTwoFiveSixK{}\,h at 256k. The quadratic term is each window's rows attending over everything before them on the global layer of every shard: $16.4\text{k}\times N^{2}$ flops per machine, 67\,TFLOP for the 64k prompt, executed at \CtxQuadraticGflops{}\,GFLOPS per machine. The decode term is linear in the context: the probe measures \ProbeAttnSlope{}\,ms per thousand positions per machine, all of it the CPU attention over the cached keys (the iGPU work does not depend on the context), which at 64k is 2.1\,GFLOP in \ProbeAttnSixtyFourK{}\,ms, again \ProbeAttnGflopsSixtyFourK{}\,GFLOPS. Both rates are the throughput of one core, and the telemetry agrees: during every request each machine had \CtxCoresBusy{} CPU cores busy while its iGPU was \CtxGpuBusySixtyFourK{}--\CtxGpuBusyOneK{}\% busy. The attention routine computes one row's 64 query heads sequentially; rows of a frame run in parallel (Section~\ref{sec:streaming}), but a lone stream's frame has one row, and a prompt window's rows are few next to the keys they attend over. The loop leaves most of each machine's sixteen cores unused. This identifies head- and key-block parallelism, followed by iGPU attention, as engineering directions; their speedups are not measured here. Larger prefill windows did not help: 64-row windows prefill at the same \CtxPrefillRateSmall{} tokens/s as 8-row ones at small contexts, so the fused-expert prefill path is bound by reading each window's distinct experts, not by fixed cost per window.

For a user, the deployment as measured serves contexts up to about 4k tokens interactively after a 22--76\,s wait for the first token, 8--16k with a 3--8\,min wait and 2.7--3.3 decode tokens/s, and 32k--64k only as batch work (27\,min and 1\,h\,50\,min to the first token, \CtxDecodeThirtyTwoK{} and \CtxDecodeSixtyFourK{} tokens/s). Longer contexts fit in memory up to 512k but are not served at useful speed with the attention kernel as it stands.

\section{Draft evaluation on deployed states}
\label{sec:drafts}

\subsection{Capturing the serving model's actual computation}
A draft head consumes internal representations whose values depend on the serving model's quantization and execution path. We evaluate Inkling's shipped multi-token prediction (MTP) head using states generated by the distributed fleet itself. The capture format records FP32 residuals together with positions, input token IDs and actual emitted token IDs. FP32 accommodates the observed pre-final-normalization range, which includes a sampled magnitude of 4,272,968.

The corpus consists of 36 sequences, three from each of twelve task families, with 160 generated positions per sequence. The 5,760 emitted tokens provide 5,724 first-draft prediction targets. All captured residuals pass finite-value checks, prompt token IDs match the rendered inputs, and decoded full responses match the completion API output. Actual emitted IDs define the evaluation labels, preserving the numerical decisions made by the deployed model.

Each of the eight MTP modules maintains its own attention and convolution history. Replay can supply true-token context or self-fed context while preserving the module-specific temporal state. This gives a controlled basis for studying the draft's representation, vocabulary and numerical format independently of request-serving timing.

\subsection{Separating vocabulary and weight-format effects}
On the same captured fleet states, the original full head yields \MTPFleet\% first-draft agreement. Restricting the original head to a prefix of 65,536 vocabulary entries yields 64.54\%. Applying INT4 feed-forward and INT8 projection/head weight grids at that vocabulary yields \MTPDeployment\%.

\begin{table}[ht]
\centering\small
\begin{tabular}{lr}
\toprule
Draft calculation on captured fleet states & First-draft agreement\\
\midrule
Original weights, full head & \MTPFleet\%\\
Original weights, 65,536-token head & 64.54\%\\
INT4/INT8 weight grids, 65,536-token head & \MTPDeployment\%\\
\bottomrule
\end{tabular}
\caption{Matched-state draft evaluation over 5,724 prediction targets. The weight-grid calculation uses FP32 arithmetic offline; the table measures agreement with emitted fleet IDs, not live speculative throughput.}
\label{tab:drafts}
\end{table}

Vocabulary selection accounts for 2.271 percentage points of the agreement difference, and additional weight quantization at the fixed vocabulary accounts for 0.175 percentage points. This decomposition identifies vocabulary size as the larger contributor in this specific deployment-format comparison. It provides a concrete basis for treating vocabulary footprint and draft-weight precision as separate design choices.

\begin{figure}[ht]
\centering\includegraphics[width=\linewidth]{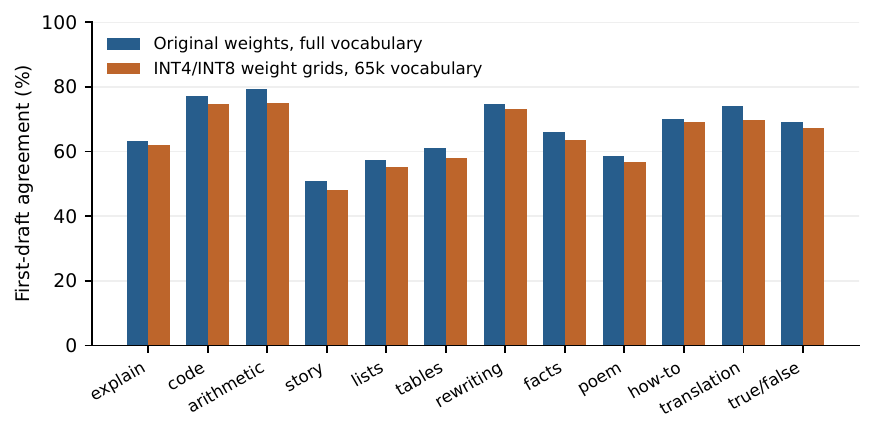}
\caption{First-draft agreement by task family on captured fleet states. Each family contains three sequences. The two calculations share the same prediction targets and differ in draft weights and vocabulary.}
\label{fig:mtp}
\end{figure}

The deployment-grid calculation records agreement of 0.746 for code, 0.748 for arithmetic and 0.482 for story. Across the eight-module chain, the reported expected number of accepted drafts is 2.128 with true-token context and 2.007 with self-fed context. These are descriptive offline measurements on the evaluated corpus. They complement the live phrase/model proposer results by exposing the draft head's behavior on the distributed model's actual states.

\subsection{Contribution to draft evaluation}
MTP and feature-based drafting are established approaches~\cite{mtp,eagle}. MoE-specific speculative systems further connect acceptance, expert activation and verification cost~\cite{moesd,moespec,specmoe}. Our contribution is an evaluation method and empirical decomposition tied to the numerical path of a distributed, compressed Inkling deployment: capture its residuals and emitted IDs, replay the draft's full temporal state, and compare vocabulary and weight formats on those same labels. This connects model-level draft analysis to the exact serving configuration being studied.

\section{Relationship to prior work}
\label{sec:related}

\paragraph{Client-cluster inference.}
Petals, TPI-LLM, prima.cpp and exo demonstrate inference across consumer or edge devices~\cite{petals,tpi,prima,exo}. Cascadia's earlier work develops Intel/OpenVINO pipeline shards and the broader distributed architecture~\cite{shards,architecture}. The present work extends that lineage to a resident 975B sparse decoder on eleven 64\,GB integrated-GPU machines, including compressed expert placement and dense/sparse operator unification. A reported Inkling deployment on eight DGX Sparks uses tensor parallelism, NVFP4, MTP and a different memory/interconnect configuration~\cite{sparkinkling}. The systems establish distinct deployment designs; the present evaluation uses the Intel fleet's own workloads and numerical formats.

\paragraph{Sparse execution and operator representation.}
EdgeMoE and PowerInfer exploit sparsity and memory placement on resource-constrained hardware~\cite{edgemoe,powerinfer}; Klotski, MegaScale-Infer, EC2MoE and CPU/GPU hybrid serving examine expert-aware execution and distribution~\cite{klotski,megascale,ec2moe,hybrid2026}. Cascadia combines a resident layer partition with a CPU/iGPU split inside each shard. Its model-specific exporter and routing interface build on OpenVINO's compressed MoE primitive and lowering passes~\cite{ovmoe,ovmoelowering}; its layer/row scaling applies equivalent-rescaling principles to the FP16 expert and FP32 residual boundary~\cite{smoothquant}. MoEfication and MLPMoE establish dense-to-expert decompositions~\cite{moefication,mlpmoe}. The measured contribution here is using all-active slices to map Inkling's dense blocks onto the same compressed fused operator family as its sparse blocks on Arc B390.

\paragraph{Streaming and draft evaluation.}
Orca and SARATHI/Sarathi-Serve establish iteration-level scheduling and chunked prefill~\cite{orca,sarathi,sarathiserve}. We integrate those principles with a gigabit sparse-model pipeline, short-convolution state and a direct sampled-token return path. The draft evaluation contributes captured-state measurements of the deployed model, complementing speculative decoding, MTP and MoE-specific verification research~\cite{specdecode,mtp,moesd}. The evidence supports these architectural, implementation and empirical contributions without requiring a priority claim for the underlying algebra or scheduling primitives.

\section{Conclusion}
Cascadia demonstrates resident inference of Inkling's 975B MoE decoder on eleven Panther Lake AI PCs. Its custom engine preserves Inkling's routing rules, constructs compressed graphs for OpenVINO's fused iGPU primitives, manages numerical range across FP16 and FP32, and unifies dense and sparse blocks within resident layer shards. Stream-oriented coordination connects those shards across machines. Its paired concurrency curve reaches \SurveyDecode{} aggregate decode tokens/s and \SurveyPhase{} whole-phase tokens/s at 88 streams, with \SurveyTTFT{}\,s median first-token latency at fifteen streams. Measured to 64k tokens of context, the deployment recovers the embedded code in all 19 measured answers while its prefill and decode costs follow one CPU core's attention throughput, which identifies the next engineering step. Captured-state draft evaluation further separates vocabulary and weight-format effects using labels emitted by the deployed model. These results provide an implemented architecture and measured techniques for serving large sparse models on distributed client hardware.

\paragraph{Artifact.}
The paper source, figures, derived tables and supporting evidence are available to reviewers in the private repository \url{https://github.com/labscommunity/cascadia-inkling-panther-lake-paper}. The evidence is pinned to Cascadia commit \path{eb7fb6381e62c1a33ec7038422bf6e8c52c77416}. \path{reproduction/CLAIMS.md} maps the paper's contributions and numerical statements to source records; the analysis scripts regenerate the reported tables and figures.

\bibliographystyle{plainnat}
\bibliography{references}
\end{document}